\pdfoutput=1

\documentclass[11pt]{article}

\usepackage{emnlp2021}

\usepackage{times}
\usepackage{latexsym}
\usepackage{CJKutf8}
\usepackage{amsmath}
\usepackage{verbatim}
\usepackage{booktabs}
\usepackage{bm}
\usepackage{amssymb}
\usepackage{pifont}
\usepackage{stfloats}
\usepackage{multicol}
\usepackage{float}
\usepackage{graphicx}
\usepackage{subfigure}
\usepackage{multirow}
\usepackage{array}
\newcommand{\PreserveBackslash}[1]{\let\temp=\\#1\let\\=\temp}
\newcolumntype{C}[1]{>{\PreserveBackslash\centering}p{#1}}
\newcolumntype{R}[1]{>{\PreserveBackslash\raggedleft}p{#1}}
\newcolumntype{L}[1]{>{\PreserveBackslash\raggedright}p{#1}}
\usepackage{diagbox}
\usepackage{booktabs}
\usepackage[ruled,linesnumbered,lined,commentsnumbered]{algorithm2e}
\usepackage{mathtools}

\usepackage[T1]{fontenc}

\usepackage[utf8]{inputenc}

\usepackage{microtype}

\title{Universal Simultaneous Machine Translation \\ with Mixture-of-Experts Wait-k Policy}

\author{Shaolei Zhang \textsuperscript{\rm 1,2},
    Yang Feng \textsuperscript{\rm 1,2}\thanks{ $\;\;$Corresponding author: Yang Feng. $\;\;\;\;\;\;\;\;\;\;\;\;\;\;\;\;$ Code is available at: \url{https://github.com/ictnlp/MoE-Waitk}} \\
        \textsuperscript{\rm 1}{Key Laboratory of Intelligent Information Processing} \\ Institute of Computing Technology, Chinese Academy of Sciences (ICT/CAS) \\
    { \textsuperscript{\rm 2} {University of Chinese Academy of Sciences, Beijing, China}} \\
     \texttt{\{zhangshaolei20z, fengyang\}@ict.ac.cn}  }

\begin{document}
\maketitle
\begin{abstract}

Simultaneous machine translation (SiMT) generates translation before reading the entire source sentence and hence it has to trade off between translation quality and latency. To fulfill the requirements of different translation quality and latency in practical applications, the previous methods usually need to train multiple SiMT models for different latency levels, resulting in large computational costs. In this paper, we propose a universal SiMT model with \emph{Mixture-of-Experts Wait-k Policy} to achieve the best translation quality under arbitrary latency with only one trained model. Specifically, our method employs multi-head attention to accomplish the mixture of experts where each head is treated as a wait-k expert with its own waiting words number, and given a test latency and source inputs, the weights of the experts are accordingly adjusted to produce the best translation. Experiments on three datasets show that our method outperforms all the strong baselines under different latency, including the state-of-the-art adaptive policy.
\end{abstract}

\section{Introduction}


Simultaneous machine translation (SiMT) \cite{Cho2016,gu-etal-2017-learning,ma-etal-2019-stacl,Arivazhagan2019} begins outputting translation before reading the entire source sentence and hence has a lower latency compared to full-sentence machine translation. In practical applications, SiMT usually has to fulfill the requirements with different levels of latency.  For example, a live broadcast requires a lower latency to provide smooth translation while a formal conference focuses on translation quality and allows for a slightly higher latency. Therefore, an excellent SiMT model should be able to maintain high translation quality under different latency levels.

\begin{figure}[t]
\centering
\includegraphics[width=2.65in]{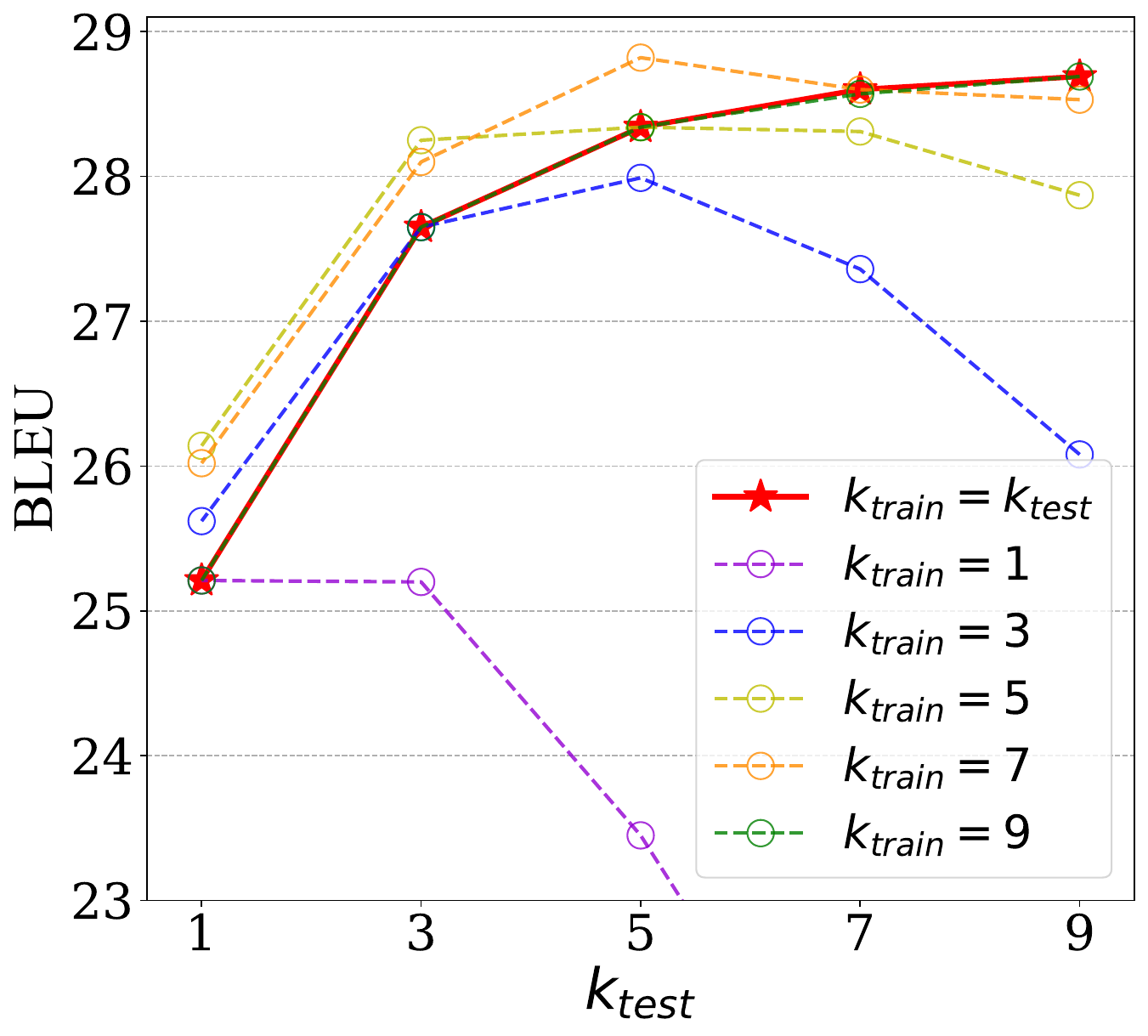}
\caption{Performance of wait-k models with different $k_{train}$ v.s. $k_{test}$ on IWSLT15 En$\rightarrow$Vi. $k_{train}$ and $k_{test}$ mean the number of source tokens to wait before translating during training and testing, respectively.}
\label{ktrain_ktest}
\end{figure}


However, the existing SiMT methods, which usually employ fixed or adaptive policy, cannot achieve the best translation performance under different latency with only one model \cite{ma-etal-2019-stacl,Ma2019a}. With fixed policy, e.g., wait-k policy \cite{ma-etal-2019-stacl}, the SiMT model has to wait for a fixed number of source words to be fed and then read one source word and output one target word alternately. In wait-k policy, the number of words to wait for can be different during training and testing, denoted as $k_{train}$ and $k_{test}$ respectively, and the latency is determined by $k_{test}$. Figure \ref{ktrain_ktest} gives the performance of the model trained with $k_{train}$ under different $k_{test}$, and the results show that under different $k_{test}$ the SiMT model with the best performance corresponds to different $k_{train}$. As a result, multiple models should be maintained for the best performance under different latency. With adaptive policy, the SiMT model dynamically adjusts the waiting of source tokens for better translation by directly involving the latency in the loss function \cite {Arivazhagan2019, Ma2019a}. Although the adaptive policy achieves the state-of-the-art performance on the open datasets, multiple models need to be trained for different latency as the change of model latency is realized by the alteration of the loss function during training. Therefore, to perform SiMT under different latency, both kinds of methods require training multiple models for different latency, leading to large costs. 

Under these grounds, we propose a universal simultaneous machine translation model which can self-adapt to different latency, so that only one model is trained for different latency. To this end, we propose a \textit{Mixture-of-Experts Wait-k Policy} (\textit{MoE wait-k policy}) for SiMT where each expert employs the wait-k policy with its own number of waiting source words. For the mixture of experts, we can consider that different experts correspond to different parameter subspaces \cite{future-guided}, and fortunately the multi-head attention is designed to explore different subspaces with different heads \cite{NIPS2017_7181}. Therefore, we employ multi-head attention as the implementation manner of MoE by assigning different heads with different waiting words number (wait-1,wait-3,wait-5,$\cdots$ ). Then, the outputs of different heads (aka experts) are combined with different weights, which are dynamically adjusted to achieve the best translation under different latency.

Experiments on IWSLT15 En$\rightarrow$Vi, WMT16 En$\rightarrow$Ro and WMT15 De$\rightarrow$En show that although with only a universal SiMT model, our method can outperform strong baselines under all latency, including the state-of-the-art adaptive policy. Further analyses show the promising improvements of our method on efficiency and robustness.



\section{Background}
Our method is based on mixture-of-experts approach, multi-head attention and wait-k policy, so we first briefly introduce them respectively.
\subsection{Mixture of Experts}
Mixture of experts (MoE) \cite{10.1162/neco.1991.3.1.79,DeepExperts,shazeer2017outrageously,peng-etal-2020-mixture} is an ensemble learning approach that jointly trains a set of expert modules and mixes their outputs with various weights:
\begin{equation}
    \mathrm{MoE}=\sum_{i=1}^{n} G_{i} \cdot \mathbf{E}_{i} 
\end{equation}
where $n$ is the number of experts, $\mathbf{E}_{i}$ and $G_{i}$ are the outputs and weight of the $i^{th}$ expert, respectively.

\subsection{Multi-head Attention}
Multi-head attention is the key component of the state-of-the-art Transformer architecture \cite{NIPS2017_7181}, which allows the model to jointly attend to information from different representation subspaces. Multi-head attention contains $h$ attention heads, where each head independently calculates its outputs between queries, keys and values through scaled dot-product attention. Since our method and wait-k policy are applied to cross-attention, the following formal expressions are all based on cross-attention, where the queries come from the $t^{th}$ decoder hidden state $\mathbf{S}_{t}$, and the keys and values come from the encoder outputs $\mathbf{Z}$. Thus, the outputs $\widetilde{\mathbf{H}}_{i}^{t}$ of the $i^{th}$ head when decoding the $t^{th}$ target token is calculated as:
\begin{equation}
\begin{aligned}
    \widetilde{\mathbf{H}}_{i}^{t}&=f_{att}\left ( \mathbf{S}_{t}, \mathbf{Z}, \mathbf{Z};\bm{\theta} _{i} \right ) \\
    &=\mathrm{softmax}\!\left (  \frac{\mathbf{S}_{t}\!\mathbf{W}_{i}^{Q} \left (  \mathbf{Z}\!\mathbf{W} _{i}^{K}\right )^{\top } }{\sqrt{d_{k}}}\right )\! \mathbf{Z}\!\mathbf{W}_{i}^{V}
\end{aligned}
\end{equation}
where $f_{att}\left (\cdot  ;\bm{\theta} _{i} \right )$ represents dot-product attention of the $i^{th}$ head, $\mathbf{W}_{i}^{Q}$, $\mathbf{W}_{i}^{K}$ and $\mathbf{W}_{i}^{V}$ are learned projection matrices, $\sqrt{d_{k}}$ is the dimension of keys. Then, the outputs of $h$ heads are concatenated and fed through a learned output matrix $\mathbf{W}^{O}$ to calculate the context vector $\mathbf{C}_{t}$:
\begin{equation}
    \mathbf{C}_{t}\!=\!\mathrm{MultiHead}\left ( \mathbf{S}_{t}, \mathbf{Z}, \mathbf{Z} \right )\!=\!\left [ \widetilde{\mathbf{H}}_{1}^{t},\!\cdots \!,\widetilde{\mathbf{H}}_{h}^{t} \right ]\mathbf{W}^{O}
    \label{eq2}
\end{equation}

\subsection{Wait-k Policy}
Wait-k policy \cite{ma-etal-2019-stacl} refers to first waiting for $k$ source tokens and then reading and writing one token alternately. Since $k$ is input from the outside of the model, we call $k$ the \textit{external lagging}. We define $g\left ( t \right )$ as a monotonic non-decreasing function of $t$, which represents the number of source tokens read in when generating the $t^{th}$ target token. In particular, for wait-k policy, given external lagging $k$, $g\left ( t;k \right )$ is calculated as:
\begin{gather}
g\left ( t;k \right )\!=\!\min\!\left \{ k\!+\!t\!-\!1, \left | \mathbf{Z} \right | \right \},\;t\!=\!1,2,\!\cdots
\end{gather}

In the wait-k policy, the source tokens processed by the encoder are limited to the first $g\left ( t;k \right )$ tokens when generating the $t^{th}$ target token. Thus, each head outputs in the cross-attention is calculated as:
\begin{gather}
    \mathbf{H}_{i}^{t}=f_{att}\left ( \mathbf{S}_{t}, \mathbf{Z}_{\leq g\left ( t;k \right )}, \mathbf{Z}_{\leq g\left ( t;k \right )};\;\bm{\theta} _{i} \right ) \label{eq6}
\end{gather}
where $\mathbf{Z}_{\leq g\left ( t;k \right )}$ represents the encoder outputs when the first $g\left ( t;k \right )$ source tokens are read in.

The standard wait-k policy \cite{ma-etal-2019-stacl} trains a set of SiMT models, where each model is trained through a fixed wait-$k_{train}$ and tested with corresponding wait-$k_{test}$ ($k_{test}\!=\!k_{train}$). \citet{multipath} proposed multipath training, which uniformly samples $k_{train}$ in each batch during training. However, training with both $k_{train}\!=\!1$ and $k_{train}\!=\!\infty$ definitely make the model parameters confused between different subspace distributions.

\section{The Proposed Method}
In this section, we first view multi-head attention from the perspective of the mixture of experts, and then introduce our method based on it.

\subsection{Multi-head Attention from MoE View}
Multi-head attention can be interpreted from the perspective of the mixture of experts \cite{peng-etal-2020-mixture}, where each head acts as an expert. Thus, Eq.(\ref{eq2}) can be rewritten as:
\begin{align}
    \mathbf{C}_{t}&=\mathrm{MultiHead}\left ( \mathbf{S}_{t},\!\mathbf{Z},\!\mathbf{Z} \right )\!=\!\left [ \widetilde{\mathbf{H}}_{1}^{t},\!\cdots \!,\widetilde{\mathbf{H}}_{h}^{t} \right ]\mathbf{W}^{O} \notag \\
    &=\left [ \widetilde{\mathbf{H}}_{1}^{t},\cdots,\widetilde{\mathbf{H}}_{h}^{t} \right ]\left [ \mathbf{W}^{O}_{1},\cdots,\mathbf{W}^{O}_{h} \right ]^{\top}  \notag\\
    &=\sum_{i=1}^{h}\widetilde{\mathbf{H}}_{i}^{t}\mathbf{W}^{O}_{i}\;\;=\;\sum_{i=1}^{h}\frac{1}{h}\cdot h\widetilde{\mathbf{H}}_{i}^{t}\mathbf{W}^{O}_{i}  \notag \\
    &=\sum_{i=1}^{h} \widetilde{G}_{i}^{t} \cdot \widetilde{\mathbf{E}}_{i}^{t} \\
    &\;\mathrm{where}\;\;\;\; \widetilde{G}_{i}^{t}=\frac{1}{h}\;\;\;,\;\;\;\widetilde{\mathbf{E}}_{i}^{t}=h\widetilde{\mathbf{H}}_{i}^{t}\mathbf{W}^{O}_{i}
    \label{eq7}
\end{align}
$\left [ \mathbf{W}^{O}_{1},\cdots,\mathbf{W}^{O}_{h} \right ]^{\top}$ is a row-wise block sub-matrix representation of $\mathbf{W}^{O}$. $\widetilde{\mathbf{E}}_{i}^{t}$ is the outputs of the $i^{th}$ expert at step $t$, and $\widetilde{G}_{i}^{t} \!\in \!\mathbb{R}$ is the weight of $\widetilde{\mathbf{E}}_{i}^{t}$. Therefore, multi-head attention can be regarded as a mixture of experts, where experts have the same function but different parameters ($\widetilde{\mathbf{E}}_{i}^{t}=h\widetilde{\mathbf{H}}_{i}^{t}\mathbf{W}^{O}_{i}$) and the normalized weights are equal ($\widetilde{G}_{i}^{t}=\frac{1}{h}$).

\subsection{Mixture-of-Experts Wait-k Policy}
\label{sec:method}

\begin{figure}[t]
\centering
\includegraphics[width=3.05in]{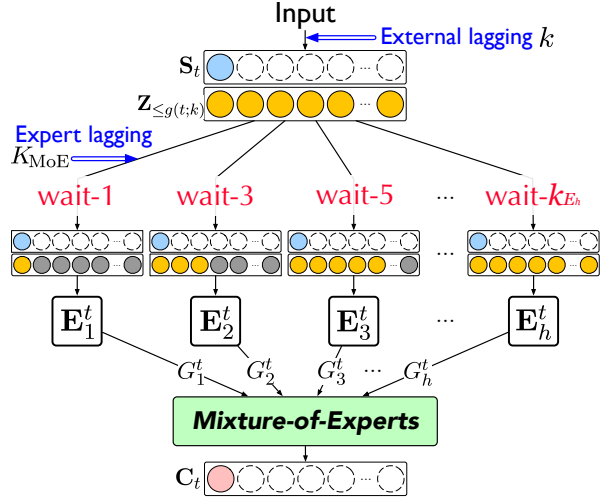}
\caption{The architecture of the mixture-of-experts wait-k policy. Each expert performs wait-k under different lagging (such as wait-1,wait-3,wait-5,$\cdots$), and then their outputs are combined with different weights.}
\label{model}
\end{figure}

To get a universal model which can perform SiMT with a high translation quality under arbitrary latency, we introduce the \textit{Mixture-of-Experts Wait-k Policy (MoE wait-k)} into SiMT to redefine the experts $\widetilde{\mathbf{E}}_{i}^{t}$ and weights $\widetilde{G}_{i}^{t}$ in multi-head attention (Eq.(\ref{eq7})). As shown in Figure \ref{model}, experts are given different functions, i.e., performing wait-k policy with different latency, and their outputs are denoted as $\left \{ \mathbf{E}_{i}^{t}\right \}_{i=1}^{h}$. Meanwhile, under the premise of normalization, the weights of experts are no longer equal but dynamically adjusted according to source input and latency requirement, denoted as $\left \{G_{i}^{t}\right \}_{i=1}^{h}$. The details are introduced following.


\subsubsection{Experts with Different Functions}
The experts in our method are divided into different functions, where each expert performs SiMT with different latency. In addition to the external lagging $k$ in standard wait-k policy, we define \textit{expert lagging} $K_{\mathrm{\!MoE}}\!=\!\left [ k_{E_{1}},\cdots, k_{E_{h}} \right ]$, where $k_{E_{i}}$ is the hyperparameter we set to represent the fixed lagging of the $i^{th}$ expert. For example, for a Transformer with 8 heads, if we set $K_{\mathrm{\!MoE}}\!=\!\left [ 1,3,5,7,9,11,13,15 \right ]$, then each expert corresponds to one head and 8 experts concurrently perform wait-1, wait-3, wait-5,$\cdots $, wait-15 respectively. Specifically, given $K_{\mathrm{\!MoE}}$, the outputs $\mathbf{H}_{i}^{t}$ of the $i^{th}$ head at step $t$ is calculated as:
\begin{align}
    \mathbf{H}_{i}^{t}=f_{att}&\left ( \mathbf{S}_{t},\; \mathbf{Z}_{\leq \min(g\left ( t;k_{E_{i}} \right ),g\left ( t;k \right ))}, \notag\right.
    \\
    &\;\;\;\phantom{=\;\;}
    \left.
    \mathbf{Z}_{\leq \min(g\left ( t;k_{E_{i}} \right ),g\left ( t;k \right ))} ;\;\bm{\theta} _{i} \right ) \label{eq8}
\end{align}
where $ g\left ( t;k_{E_{i}} \right )$ is the number of source tokens processed by the $i^{th}$ expert at step $t$ and $g\left ( t;k \right )$ is the number of all available source tokens read in at step $t$. During training, $k$ is uniformly sampled in each batch with multipath training \cite{multipath}. During testing, $k$ is the input test lagging.

Then, the outputs $\mathbf{E}_{i}^{t}$ of the $i^{th}$ expert when generating $t^{th}$ target token is calculated as:
\begin{equation}
    \mathbf{E}_{i}^{t}=h\mathbf{H}_{i}^{t}\mathbf{W}^{O}_{i}\label{eq9}
\end{equation}

\subsubsection{Dynamic Weights for Experts}
Each expert has a clear division of labor through expert lagging $K_{\mathrm{\!MoE}}$. Then for different input and latency, we dynamically weight each expert with the predicted $\left \{G_{i}^{t}\right \}_{i=1}^{h}$, where $G_{i}^{t}\!\in \!\mathbb{R}$ can be considered as the confidence of expert outputs $\mathbf{E}_{i}^{t}$.
The factor to predict $G_{i}^{t}$ consists of two components:
\begin{itemize}
\setlength{\itemsep}{0pt}
\setlength{\parsep}{0pt}
\setlength{\parskip}{0pt}
    \item $e_{i}^{t}$: The average cross-attention scores in the $i^{th}$ expert at step $t$, which are averaged over all source tokens read in \cite{Zheng2019b}.
    \item $k$: External lagging $k$ in Eq.(\ref{eq8}).
\end{itemize}

At step $t$, all $e_{i}^{t}$ and $k$ are concatenated and fed through the multi-layer perceptron (MLP) to predict the confidence score $\beta _{i}^{t}$ of the $i^{th}$ expert, which are then normalized to calculate the weight $G_{i}^{t}$:
\begin{align}
    \beta _{i}^{t}=\mathrm{tanh}&\left (\; [e_{1}^{t};\cdots;e_{h}^{t};k]\mathbf{W}_{i}+b_{i} \;\right ) \label{eq10} \\
    G_{i}^{t}&=\frac{\exp(\beta _{i}^{t})}{\sum_{l=1}^{h}\exp(\beta _{l}^{t})} \label{eq11}
\end{align}
where $\mathbf{W}_{i}$ and $b_{i}$ are parameters of MLP to predict $G_{i}^{t}$. Given expert outputs $\left \{ \mathbf{E}_{i}^{t}\right \}_{i=1}^{h}$ and weights $\left \{G_{i}^{t}\right \}_{i=1}^{h}$, the context vector $\mathbf{C}_{t}$ is calculatas:
\begin{equation}
    \mathbf{C}_{t}=\sum_{i=1}^{h} G_{i}^{t} \cdot \mathbf{E}_{i}^{t} \label{eq12}
\end{equation}

The algorithm details of proposMoE wait-k policy are shown in Algorithm \ref{algorithm}. At decoding step $t$, each expert performs the wait-k policy with different latency according to the expert lagging $K_{\mathrm{\!MoE}}$, and then the expert outputs are dynamically weighted to calculate the context vector $\mathbf{C}_{t}$.


\subsubsection{Training Method}
\label{training}
We apply a two-stage training, both of which apply multipath training \cite{multipath}, i.e., randomly sampling $k$ ($k$ in Eq.(\ref{eq8})) in every batch during training. \textit{First-stage}: Fix the weights $G_{i}^{t}$ equal to $\frac{1}{h}$ and pre-train expert parameters. \textit{Second-stage}: jointly fine-tune the parameters of experts and their weights. In the inference time, the universal model is tested with arbitrary latency (test lagging). In Sec.\ref{experiments}, we compare the proposed two-stage training method with the one-stage training method which directly trains the parameters of experts and their weights together.

We tried the block coordinate descent (BCD) training \cite{peng-etal-2020-mixture} which is proposed to train the experts in the same function, but it is not suitable for our method, as the experts in MoE wait-k have already assigned different functions. Therefore, our method can be stably trained through back-propagation directly.

\begin{algorithm}[t]
\caption{MoE Wait-k Policy}\label{algorithm}
  \SetKwData{Training}{Training}\SetKwData{This}{this}\SetKwData{Up}{up}
  \SetKwFunction{getInputTestLagging}{getInputTestLagging}\SetKwFunction{SampleFrom}{SampleFrom}
  \SetKwInOut{Input}{Input}\SetKwInOut{Output}{Output}
  \Input{Encoder output $\mathbf{Z}$ (incomplete), \\
  Decoder hidden state $\mathbf{S}_{t}$, \\
  Expert lagging $K_{\mathrm{\!MoE}}$,\\
  Test lagging $k_{test}$ (only in testing)}
  \Output{Context vector $\mathbf{C}_{t}$}
  \BlankLine
  \eIf(\tcp*[f]{In training}){$\mathrm{is\_Training}$}{
      $k$ $\leftarrow$ $\mathrm{Sample\;from(\;}$$[\;1,2,\cdots ,\left | \mathbf{Z} \right |\;]$$\mathrm{\;)}$
    }
    (\tcp*[f]{In testing}){
      $k$ $\leftarrow$ $k_{test}$
    }
    \BlankLine
    \For{$k_{E_{i}}$ $\mathrm{in}$ $K_{\mathrm{\!MoE}}$}{
    calculate $\mathbf{Z}_{\leq \min(g\left ( t;k_{E_{i}} \right ),g\left ( t;k \right ))} $
    }
    \BlankLine
    \For{$i$ $\leftarrow$ $1$ $\mathrm{to}$ $h$}{
    calculate $\mathbf{E}_{i}^{t}$ according to Eq.(\ref{eq8}, \ref{eq9})\\
    calculate $\mathbf{G}_{i}^{t}$ according to Eq.(\ref{eq10}, \ref{eq11})
    }
    calculate $\mathbf{C}_{t}$ according to Eq.(\ref{eq12})\\
    \BlankLine
    \textbf{Return} $\mathbf{C}_{t}$
\end{algorithm}

\section{Related Work}
\textbf{Mixture of experts} MoE was first proposed in multi-task learning \cite{10.1162/neco.1991.3.1.79,10.1145/1015330.1015432,Liu_2018_ECCV,10.1145/3219819.3220007,Dutt2018CoupledEO}. Recently, \citet{shazeer2017outrageously} applied MoE in sequence learning. Some work \cite{he-etal-2018-sequence, pmlr-v97-shen19c,cho-etal-2019-mixture} applied MoE in diversity generation. \citet{peng-etal-2020-mixture} applied MoE in MT and combined $h-1$ heads in Transformer as an expert.

Previous works always applied MoE for diversity. Our method makes the experts more regular in parameter space, which provides a method to improves the translation quality with MoE.

\textbf{SiMT} Early read / write policies in SiMT used segmented translation \cite{bangalore-etal-2012-real,Cho2016,siahbani-etal-2018-simultaneous}. \citet{grissom-ii-etal-2014-dont} predicted the final verb in SiMT. \citet {gu-etal-2017-learning} trained a read / write agent with reinforcement learning. \citet {Alinejad2019} added a predict operation based on \citet{gu-etal-2017-learning}.

Recent read / write policies fall into two categories: fixed and adaptive. For the fixed policy, \citet {dalvi-etal-2018-incremental} proposed STATIC-RW, and \citet {ma-etal-2019-stacl} proposed wait-k policy, which always generates target $k$ tokens lagging behind the source. \citet{multipath} enhanced wait-k policy by sampling different $k$ during training. \citet{han-etal-2020-end} applied meta-learning in wait-k. \citet{future-guided} proposed future-guided training for wait-k policy. \citet{zhang-feng-2021-icts} proposed a char-level wait-k policy. For the adaptive policy, \citet{Zheng2019b} trained an agent with gold read / write sequence. \citet {Zheng2019a} added a ``delay'' token $\left \{ \varepsilon  \right \}$ to read. \citet {Arivazhagan2019} proposed MILk, which used a Bernoulli variable to determine writing. \citet {Ma2019a} proposed MMA, which is the implementation of MILk on the Transformer. \citet {zheng-etal-2020-simultaneous} ensembled multiple wait-k models to develop a adaptive policy. \citet{zhang-zhang-2020-dynamic} and \citet{zhang-etal-2020-learning-adaptive} proposed adaptive segmentation policies. \citet{bahar-etal-2020-start} and \citet{wilken-etal-2020-neural} proposed alignment-based chunking policy. 

A common weakness of the previous methods is that they all train separate models for different latency. Our method only needs a universal model to complete SiMT under all latency, and meanwhile achieve better translation quality.

\section{Experiments}
\label{experiments}

\subsection{Datasets}
We evaluated our method on the following three datasets, the scale of which is from small to large.

\textbf{IWSLT15\footnote{\url{nlp.stanford.edu/projects/nmt/}} English$\rightarrow $Vietnamese (En-Vi)} (133K pairs) \cite{iwslt2015} We use TED tst2012 (1553 pairs) as the validation set and TED tst2013 (1268 pairs) as the test set. Following \citet{LinearTime} and \citet{Ma2019a}, we replace tokens that the frequency less than 5 by $\left \langle unk \right \rangle$. After replacement, the vocabulary sizes are 17K and 7.7K for English and Vietnamese, respectively.

\textbf{WMT16\footnote{\url{www.statmt.org/wmt16/}} English$\rightarrow $Romanian (En-Ro)} (0.6M pairs) \cite{lee-etal-2018-deterministic} We use news-dev2016 (1999 pairs) as the validation set and news-test2016 (1999 pairs) as the test set. 

\textbf{WMT15\footnote{\url{www.statmt.org/wmt15/}} German$\rightarrow$English (De-En)} (4.5M pairs) Following the setting from \citet{ma-etal-2019-stacl} and \citet{Ma2019a}, we use newstest2013 (3000 pairs) as the validation set and newstest2015 (2169 pairs) as the test set.

For En-Ro and De-En, BPE \cite{sennrich-etal-2016-neural} is applied with 32K merge operations and the vocabulary is shared across languages.

\subsection{System Settings}
We conducted experiments on following systems.

{\bf {Offline}} Conventional Transformer \cite{NIPS2017_7181} model for full-sentence translation, decoding with greedy search.

{\bf {Standard Wait-k}} Standard wait-k policy proposed by \citet{ma-etal-2019-stacl}. When evaluating with the test lagging $k_{test}$, we apply the result from the model trained with $k_{train}$, where $k_{train}\!=\!k_{test}$.

{\bf {Optimal Wait-k}} An optimal variation of standard wait-k. When decoding with $k_{test}$, we traverse all models trained with different $k_{train}$ and apply the optimal result among them. For example, if the best result when testing with wait-1 ($k_{test}\!=\!1$) comes from the model trained by wait-5 ($k_{train}\!=\!5$), we apply this optimal result. `Optimal Wait-k' selects the best result according to the reference, so it can be considered as an oracle.

{\bf {Multipath Wait-k}} An efficient training method for wait-k policy \cite{multipath}. In training, $k_{train}$ is no longer fixed, but randomly sampled from all possible lagging in each batch.

{\bf {MU}} A segmentation policy base on meaning units proposed by \citet{zhang-etal-2020-learning-adaptive}, which obtains comparable results with SOTA adaptive policy. At each decoding step, if a meaning unit is detected through a BERT-based classifier, `MU' feeds the received source tokens into a full-sentence MT model to generate the target token and stop until generating the $<\!EOS\!>$ token.

{\bf {MMA}\footnote{\url{github.com/pytorch/fairseq/tree/master/examples/simultaneous_translation}}} Monotonic multi-head attention (MMA) proposed by \cite{Ma2019a}, the state-of-the-art adaptive policy for SiMT, which is the implementation of `MILk' \cite {Arivazhagan2019} based on the Transformer. At each decoding step, `MMA' predicts a Bernoulli variable to decide whether to start translating or wait for the source token.

{\bf {MoE Wait-k}} A variation of our method, which directly trains the parameters of experts and their weights together in one-stage training.

{\bf {Equal-Weight MoE Wait-k}} A variation of our method. The weight of each expert is fixed to $\frac{1}{h}$.

{\bf {MoE Wait-k + FT}} Our method in Sec.\ref{sec:method}.

We compare our method with `MMA' and `MU' on De-En(Big) since they report their results on De-En with Transformer-Big.

\begin{table}[]
\begin{tabular}{c|c} \toprule[1.2pt]
\textbf{Architecture}                                                 & \textbf{Expert Lagging} $K_{\mathrm{\!MoE}}$                                                                  \\ \midrule[0.8pt]
\begin{tabular}[c]{@{}c@{}}Transformer-Small\\ (4 heads)\end{tabular} & {[}1, 6, 11, 16{]}                                                                           \\ \hline
\begin{tabular}[c]{@{}c@{}}Transformer-Base\\ (8 heads)\end{tabular}  & {[}1, 3, 5, 7, 9, 11, 13, 15{]}                                                                \\  \hline
\begin{tabular}[c]{@{}c@{}}Transformer-Big\\ (16 heads)\end{tabular}  & \begin{tabular}[c]{@{}c@{}}{[}1, 2, 3, 4, 5, 6, 7, 8,\\ 9,10,11,12,13,14,15,16{]}\end{tabular} \\\bottomrule[1.2pt]
\end{tabular}
\caption{The value of expert lagging $K_{\mathrm{\!MoE}}$ for different Transformer settings.}
\label{kme}
\end{table}

The implementation of all systems are adapted from Fairseq Library \cite{ott-etal-2019-fairseq}, and the setting is exactly the same as \citet{ma-etal-2019-stacl} and \citet{Ma2019a}. To verify that our method is effective on Transformer with different head settings, we conduct experiments on three types of Transformer, where the settings are the same as \citet{NIPS2017_7181}. For En-Vi, we apply \textbf{Transformer-Small} (4 heads). For En-Ro, we apply \textbf{Transformer-Base} (8 heads). For De-En, we apply both \textbf{Transformer-Base} and \textbf{Transformer-Big} (16 heads). Table \ref{para} reports the parameters of different SiMT systems on De-En(Big). To perform SiMT under different latency, both `Standard Wait-k', `Optimal Wait-k' and `MMA' require multiple models, while `Multipath Wait-k', `MU' and `MoE Wait-k' only need one trained model.

Expert lagging $K_{\mathrm{\!MoE}}$ in MoE wait-k is the hyperparameter we set, which represents the lagging of each expert. We did not conduct many searches on $K_{\mathrm{\!MoE}}$, but set it to be uniformly distributed in a reasonable lagging interval, as shown in Table \ref{kme}. We will analyze the influence of different settings of $K_{\mathrm{\!MoE}}$ in our method in Sec.\ref{sec:lagging}.

We evaluate these systems with BLEU \cite{papineni-etal-2002-bleu} for translation quality and Average Lagging (AL\footnote{\url{github.com/SimulTrans-demo/STACL}.}) \cite{ma-etal-2019-stacl} for latency. Given $g\left ( t \right )$, latency metric AL is calculated as:
\begin{gather}
    \mathrm{AL}=\frac{1}{\tau }\sum_{t=1}^{\tau}g\left ( t \right )-\frac{t-1}{\left | \mathbf{y} \right |/\left | \mathbf{x} \right |}\\
\mathrm{where}\;\;\;  \tau =\underset{t}{\mathrm{argmax}}\left ( g\left ( t \right )= \left | \mathbf{x} \right |\right )
\end{gather}
where $\left | \mathbf{x} \right |$ and $\left | \mathbf{y} \right |$ are the length of the source sentence and target sentence respectively.

\begin{table}[]
\begin{tabular}{l|C{1.5cm}C{0.9cm}|r} \toprule[1.2pt]
\textbf{Systems}     & \textbf{\begin{tabular}[c]{@{}c@{}}\#Para. \\ per Model\end{tabular}} & \textbf{\begin{tabular}[c]{@{}c@{}}Model\\ Num.\end{tabular}} & \multicolumn{1}{c}{\textbf{\begin{tabular}[c]{@{}c@{}}Total \\ \#Para.\end{tabular}}} \\ \midrule[0.8pt]
\textbf{Offline}    & 209.91M                                                               & 1                                                              & 209.91M                                                                               \\ 
\textbf{Wait-k}     & 209.91M                                                               & 5                                                              & 1049.55M                                                                              \\
\textbf{Optimal}     & 209.91M                                                               & 5                                                              & 1049.55M                                                                              \\
\textbf{Mulitpath} & 209.91M                                                               & 1                                                              & 209.91M            \\
\textbf{MMA}        & 222.51M                                                               & 7                                                              & 1557.57M                                                                              \\
\textbf{MU}         & 319.91M                                                               & 1                                                              & 319.91M                                                                               \\
\textbf{MoE Wait-k} & 209.91M                                                               & 1                                                              & 209.91M            \\\bottomrule[1pt]                                                                  
\end{tabular}
\caption{The parameters of SiMT systems on De-En(Transformer-Big) in our experiments. `\#Para. per model': The parameters of a single SiMT model. `Model Num.': The number of SiMT models required to perform SiMT under multiple latency. `Total \#Para.': The total parameters of the SiMT system.}
\label{para}
\end{table}

\subsection{Main Results}

\begin{figure*}[t]
\centering
\subfigure[En-Vi, Transformer-Small]{
\includegraphics[width=1.87in]{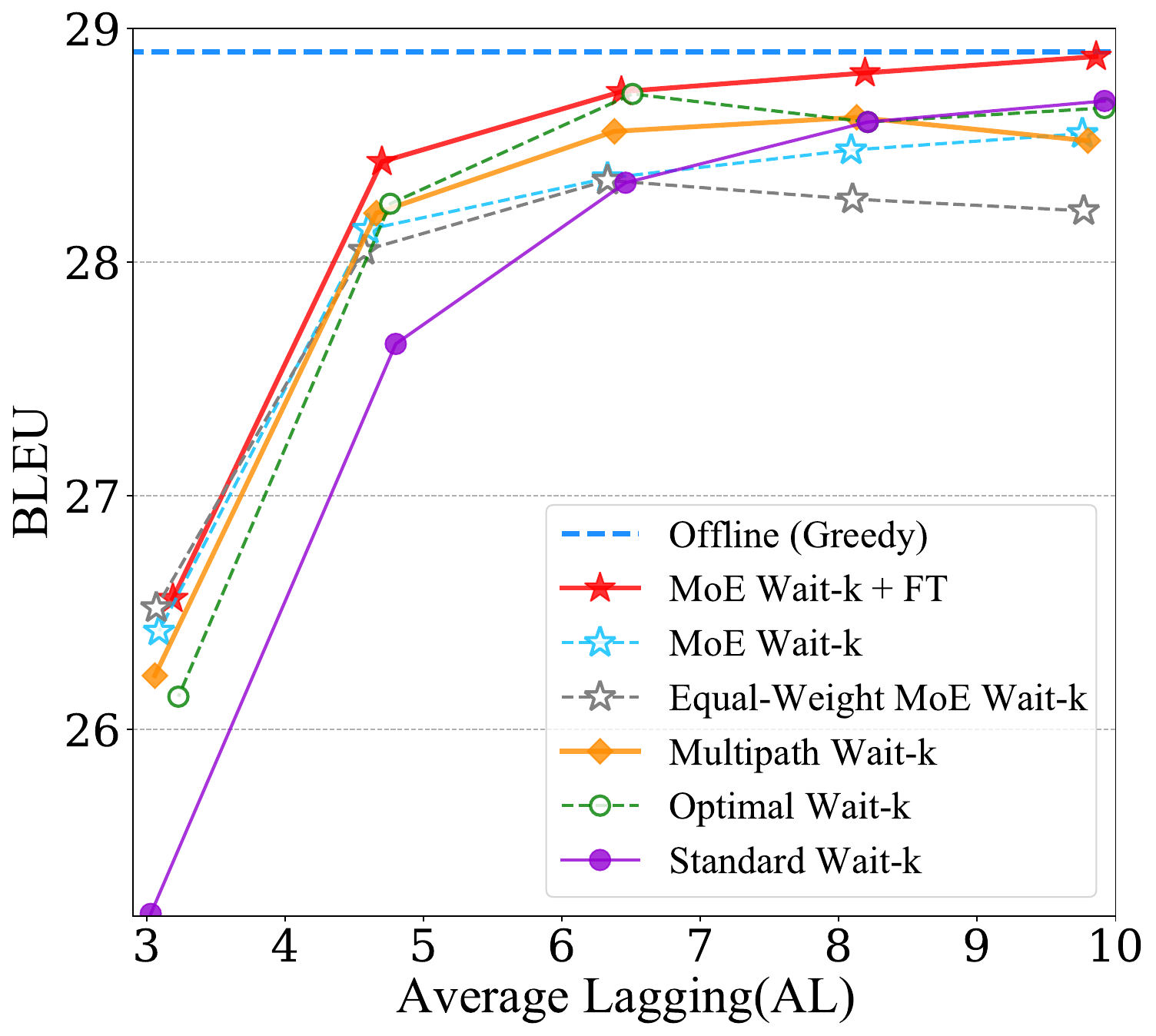}
}
\quad
\subfigure[En-Ro, Transformer-Base]{
\includegraphics[width=1.87in]{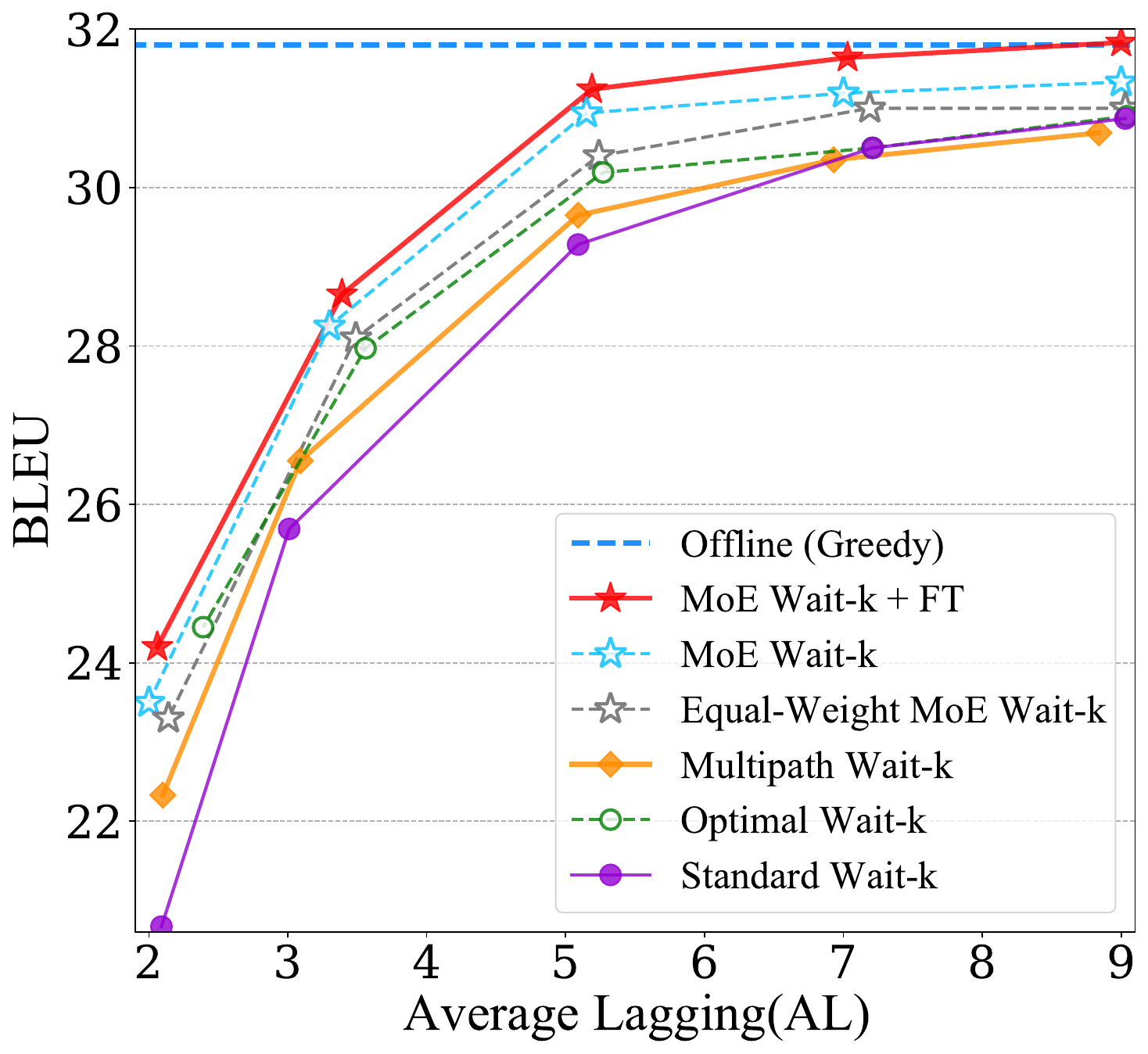}
}
\quad
\subfigure[De-En, Transformer-Base]{
\includegraphics[width=1.87in]{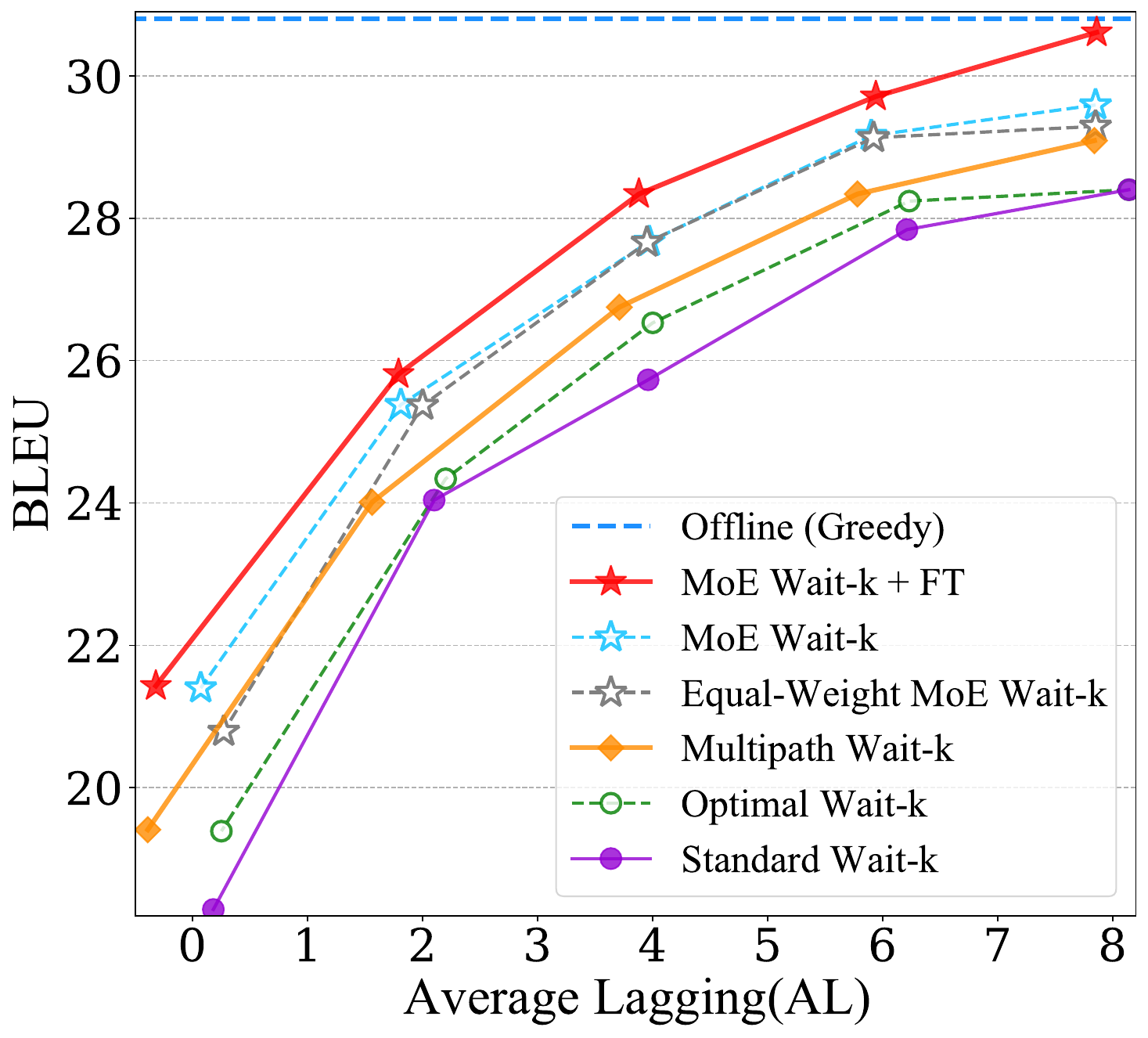}
}

\caption{Translation quality (BLEU) against latency (AL) on the En-Vi(Small), En-Ro(Base), De-En(Base). We show the result of our methods, Standard wait-k, Optimal Wait-k, Multipath Wait-k and offline model.}
\label{main}
\end{figure*}

\begin{figure}[t]
\includegraphics[width=3.0in]{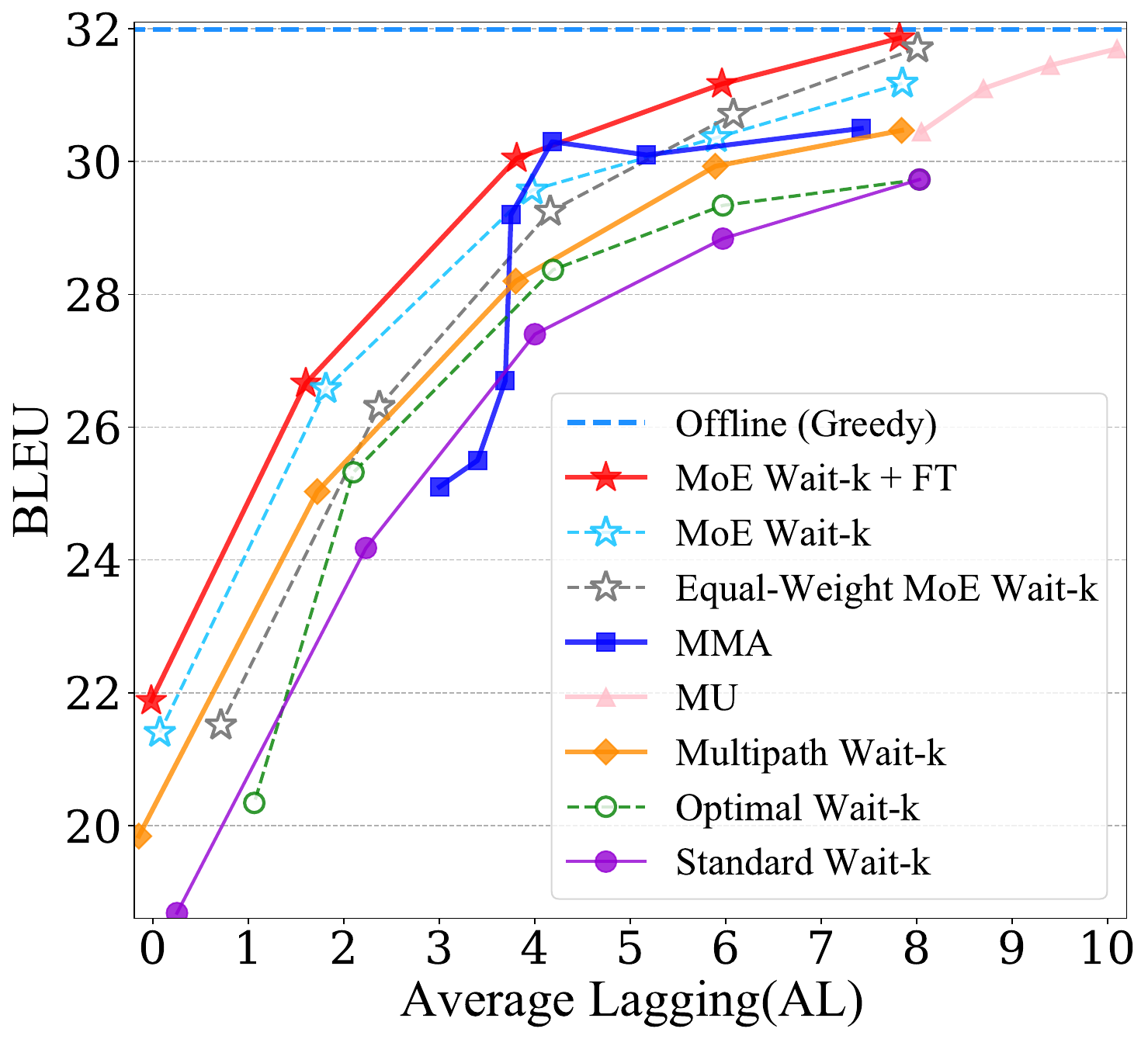}
\caption{Translation quality (BLEU) against latency (AL) on the De-En with Transformer-Big. We show the result of our methods, Standard wait-k, Optimal Wait-k, Multipath Wait-k, MU, MMA (the current SOTA adaptive policy) and offline model.}
\label{main2}
\end{figure}

Figure \ref{main} and Figure \ref{main2} show the comparison between our method and the previous methods on Transformer with the various head settings. In all settings, `MoE wait-k + FT' outperforms the previous methods under all latency. Our method improves the performance of SiMT much closer to the offline model, which almost reaches the performance of full-sentence MT when lagging 9 tokens.

Compared with `Standard Wait-k', our method improves 0.60 BLEU on En-Vi, 2.11 BLEU on En-Ro, 2.33 BLEU on De-En(Base), and 2.56 BLEU on De-En(Big), respectively (average on all latency). More importantly, our method only needs one well-trained universal model to complete SiMT under all latency, while `Standard wait-k' requires training different models for each latency. Besides, `Optimal Wait-k' traverses many models to obtain the optimal result under each latency. Our method dynamically weights experts according to the test latency, and outperforms `Optimal Wait-k' under all latency, without searching among many models.

Both our method and `Multipath Wait-k' can train a universal model, but our method avoids the mutual interference between different sampled $k$ during training. `Multipath Wait-k' often improves the translation quality under low latency, but on the contrary, the translation quality under high latency is poor \cite{elbayad-etal-2020-trac}. The reason is that sampling a slightly larger $k$ in training improves the translation quality under low latency \cite{ma-etal-2019-stacl,future-guided}, but sampling a smaller $k$ destroys the translation quality under high latency. Our method introduces expert lagging and dynamical weights, avoiding the interference caused by multipath training.

Compared with `MMA' and `MU', our method performs better. `MU' sets a threshold to perform SiMT under different latency and achieves good translation quality, but it is difficult to complete SiMT under low latency as it is a segmentation policy. As a fixed policy, our method maintains the advantage of simple training and meanwhile catches up with the adaptive policy `MMA' on translation quality, which is uplifting. Furthermore, our method only needs a universal model to perform SiMT under different latency and the test latency can be set artificially, which is impossible for the previous adaptive policy.
\subsection{Ablation Study}

We conducted ablation studies on the dynamic weights and two-stage training, as shown in Figure \ref{main} and Figure \ref{main2}. The translation quality decreases significantly when each expert is set to equal-weight. Our method dynamically adjusts the weight of each expert according to the input and test lagging, resulting in concurrently performing well under all latency. For the training methods, the two-stage training method makes the training of weights more stable, thereby improving the translation quality, especially under high latency.

\begin{table*}[]
\centering
\begin{tabular}{c|C{1.2cm}C{1.2cm}C{1.2cm}|C{1.2cm}C{1.2cm}C{1.2cm}|C{1.2cm}C{1.2cm}C{1.2cm}} \toprule[1.2pt]
\multirow{2}{*}{\bm{$k_{test}$}} & \multicolumn{3}{c|}{\textbf{EASY}}                      & \multicolumn{3}{c|}{\textbf{MIDDLE}}                    & \multicolumn{3}{c}{\textbf{HARD}}                      \\\cline{2-10}
                            & \textbf{Wait-k} & \textbf{Ours} & \bm{$\Delta$} & \textbf{Wait-k} & \textbf{Ours} & \bm{$\Delta$} & \textbf{Wait-k} & \textbf{Ours} & \bm{$\Delta$}\\ \midrule[0.8pt]
\textbf{1}                  & 19.27           & 21.79               & +2.52          & 18.70           & 21.87               & +3.17          & 16.14           & 20.04               & \textbf{+3.90} \\
\textbf{3}                  & 28.79           & 30.19               & +1.40          & 24.88           & 25.65               & +0.77          & 21.30           & 23.81               & \textbf{+2.51} \\
\textbf{5}                  & 31.15           & 33.80               & \textbf{+2.65} & 26.56           & 29.03               & +2.47          & 24.02           & 25.73               & +1.71          \\
\textbf{7}                  & 32.62           & 34.68               & \textbf{+2.06} & 28.52           & 30.42               & +1.90          & 25.65           & 27.37               & +1.72          \\
\textbf{9}                  & 32.52           & 35.08               & \textbf{+2.56} & 28.94           & 31.42               & +2.48          & 26.66           & 28.40               & +1.74    \\     
\bottomrule[1pt]
\end{tabular}
\caption{Improvement of our method on SiMT with various difficult levels, evaluated with wait-$k_{test}$. The difficult levels are divided according to the word order difference between the source sentence and the target sentence.}
\label{hard}
\end{table*}

\section{Analysis}

We conducted extensive analyses to understand the specific improvements of our method. Unless otherwise specified, all the results are reported on De-En with {Transformer}-Base(8 heads).

\subsection{Performance on Various Difficulty Levels}
The difference between the target and source word order is one of the challenges of SiMT, where many word order inversions force to start translating before reading the aligned source words. To verify the performance of our method on SiMT with various difficulty levels, we evenly divided the test set into three parts: EASY, MIDDLE and HARD. Specifically, we used \texttt{fast-align}\footnote{\url{https://github.com/clab/fast_align}} \cite{dyer-etal-2013-simple} to align the source with the target, and then calculated the number of crosses in the alignments (number of reversed word orders), which is used as a basis to divide the test set \cite{2020arXiv201011247C,future-guided}. After the division, the alignments in the EASY set are basically monotonous, and the sentence pairs in the HARD set contains at least 12 reversed word orders.

Our method outperforms the standard wait-k on all difficulty levels, especially improving 3.90 BLEU on HARD set under low latency. HARD set contains a lot of word order reversal, which is disastrous for low-latency SiMT such as testing with wait-1. The standard wait-k enables the model to gain some implicit prediction ability \cite{ma-etal-2019-stacl}, and our method further strengthens it. MoE wait-k introduces multiple experts with varying expert lagging, of which the larger expert lagging helps the model to improve the implicit prediction ability \cite{future-guided}, while the smaller expert lagging avoids learning too much future information during training and prevents the illusion caused by over-prediction \cite{2020arXiv201011247C}. With MoE wait-k, the implicit prediction ability is stronger and more stable.
\begin{figure}[t]
\includegraphics[width=3in]{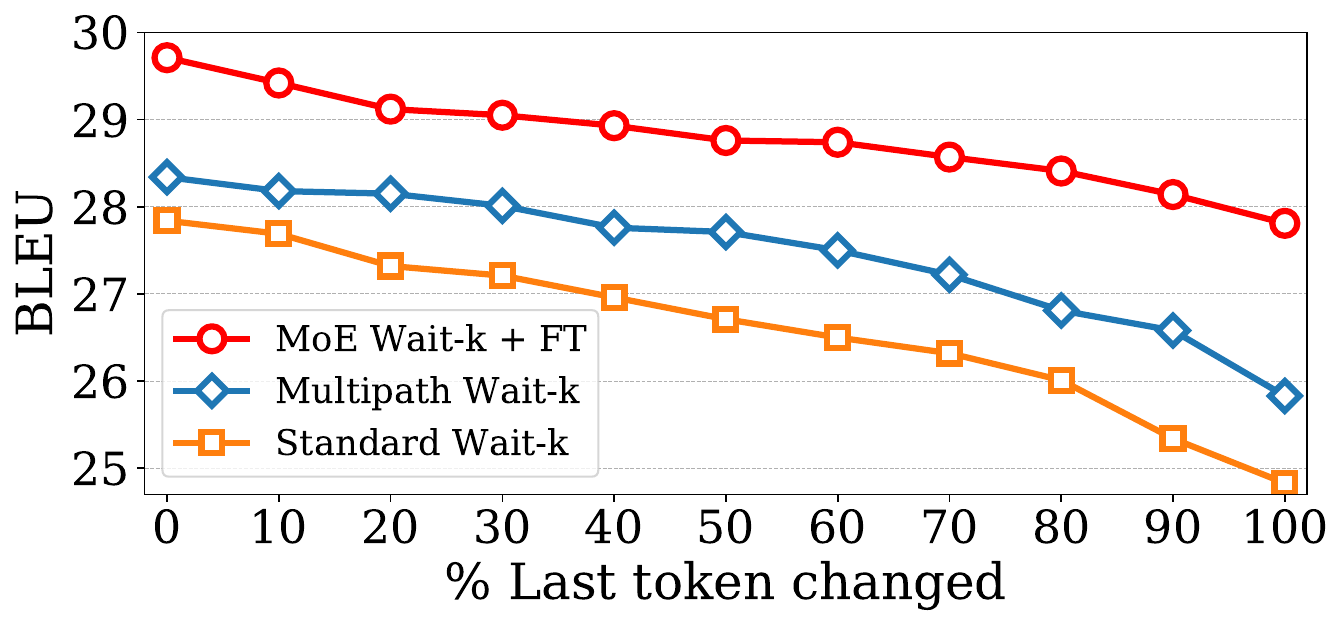}
\caption{Degradation of performance as the noise of last source token increases, evaluated with wait-7.}
\label{robust}
\end{figure}
\subsection{Improvement on Robustness}

Robustness is another major challenge for SiMT \cite{zheng-etal-2020-fluent}. SiMT is often used as a downstream task of streaming automatic speech recognition (ASR), but the results of streaming ASR are not stable, especially the last recognized source token \cite{li-etal-2020-bits,gaido-etal-2020-end,zheng-etal-2020-fluent}. In each decoding step, we randomly modified the last source token with different proportions, and the results are shown in Figure \ref{robust}.

Our method is more robust with the noisy last token, owing to multiple experts. Due to different expert lagging, the number of source tokens processed by each expert is different and some experts do not consider the last token. Thus, the noisy last token only affects some experts, while other experts would not be disturbed, giving rise to robustness.

\begin{table*}[]
\centering
\begin{tabular}{C{1.3cm}|c|C{0.8cm}C{0.8cm}C{0.8cm}C{0.8cm}C{0.8cm}C{0.8cm}C{0.8cm}C{0.8cm}|c} \toprule[1.2pt]
\multicolumn{2}{c|}{}                                                                               & $\mathbf{E_{1}}$ & $\mathbf{E_{2}}$ & $\mathbf{E_{3}}$ & $\mathbf{E_{4}}$ & $\mathbf{E_{5}}$ & $\mathbf{E_{6}}$ & $\mathbf{E_{7}}$ & $\mathbf{E_{8}}$    & \multirow{2}{*}{\begin{tabular}[c]{@{}c@{}}\textbf{Optimal}\\ \textbf{Model}\end{tabular}} \\ \cline{3-10}
\multicolumn{2}{c|}{\textbf{Expert Lagging} $K_{\mathrm{\!MoE}}$}                                                                & 1     & 3              & 5              & 7     & 9              & 11             & 13    & 15    &                                                                          \\ \midrule[0.8pt]
\multirow{5}{*}{\textbf{\begin{tabular}[c|]{@{}c@{}}Test\\ Lagging\end{tabular}}} & $k_{test}\!=\!1$ & 10.66 & \textbf{13.90} & 13.82          & 11.67 & 13.07          & 13.49          & 11.89 & 11.50 & $k_{train}\!=\!3$                                                            \\
                                                                                    & $k_{test}\!=\!3$ & $\;\;$9.83  & 12.88          & \textbf{13.75} & 11.62 & 13.70          & 13.51          & 12.55 & 12.16 & $k_{train}\!=\!5$                                                            \\
                                                                                    & $k_{test}\!=\!5$ & $\;\;$9.35  & 12.63          & 13.52          & 11.61 & \textbf{13.82} & 13.6           & 12.93 & 12.54 & $k_{train}\!=\!9$                                                            \\
                                                                                    & $k_{test}\!=\!7$ & $\;\;$8.65  & 12.55          & 12.82          & 11.58 & 14.04          & \textbf{14.10} & 13.53 & 12.73 & $k_{train}\!=\!9$                                                            \\
                                                                                    & $k_{test}\!=\!9$ & $\;\;$8.34  & 12.32          & 12.55          & 11.08 & 14.33          & \textbf{14.69} & 13.79 & 12.90 & $k_{train}\!=\!9$ \\\bottomrule[1pt]                                                     
\end{tabular}
\caption{Weight of experts under different latency, averaged on 6 decoder layers at all decoding steps. `Optimal Model': The optimal standard wait-k model under current test latency, obtained by traversing all models trained with different wait-$k_{train}$.}
\label{gate2}
\end{table*}
\begin{figure}[t]
\centering
\subfigure[Multipath Wait-k]{
\includegraphics[width=1.42in]{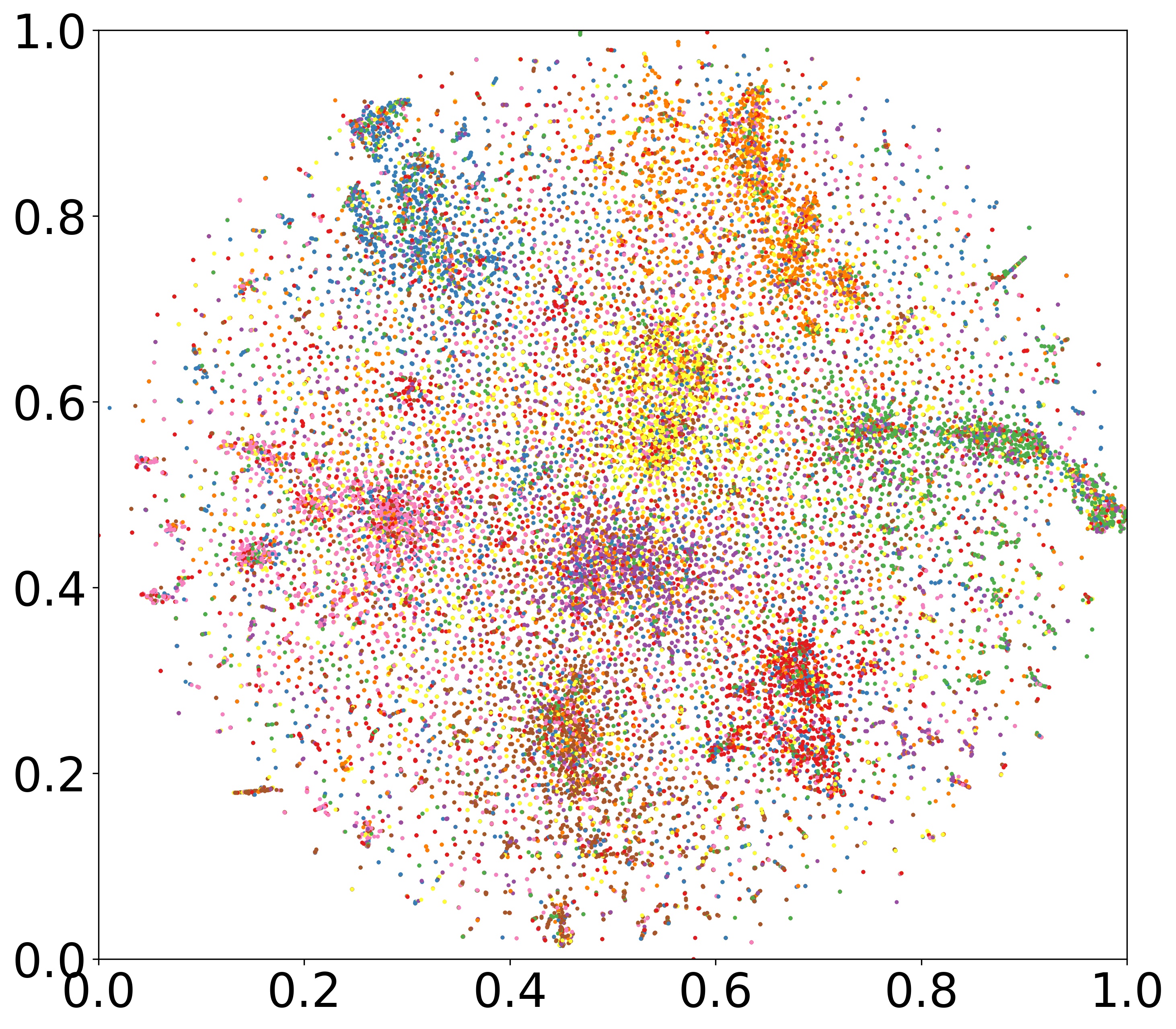}
}
\subfigure[MoE Wait-k + FT]{
\includegraphics[width=1.42in]{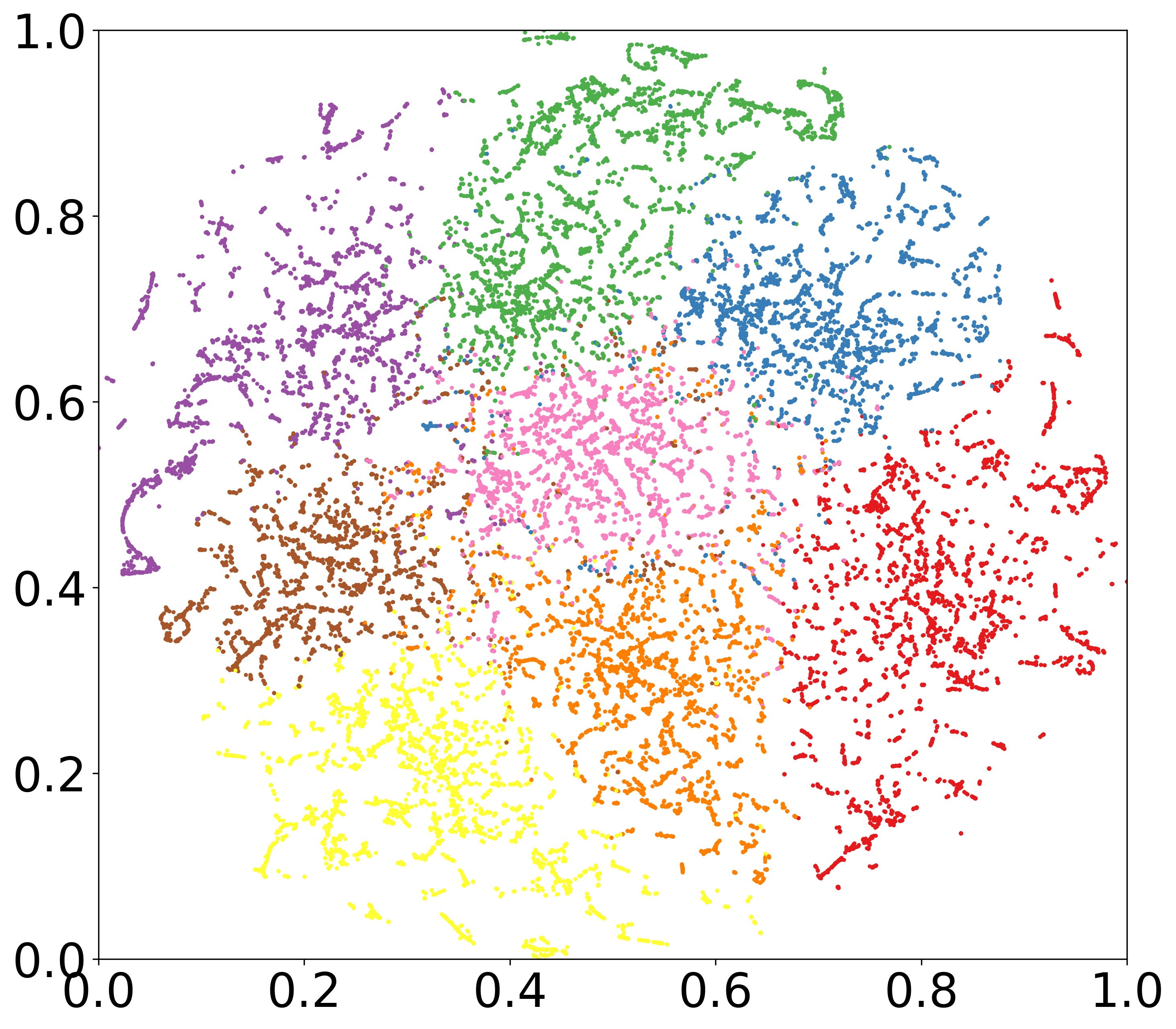}
}
\caption{Subspace distribution of expert outputs. Each color represents the outputs of an expert.}
\label{tsne}
\end{figure}
\subsection{Differentiation of Experts Distribution}
Our method clearly divides the experts into different functions and integrates the expert outputs from different subspaces for better translation. For `Multipath Wait-k' and our method, we sampled 200 cases and reduced the dimension of the expert outputs (evaluating with wait-5) with the t-Distributed Stochastic Neighbor Embedding (tSNE) technique, and shown the subspace distribution of the expert outputs in Figure \ref{tsne}.

The expert outputs in `Multipath Wait-k' have a little difference but most of them are fused together, which shows some similarities in heads. In our method, due to the clear division of labor, the expert outputs are significantly different and regular in the subspace distribution, which proves to be beneficial to translation \cite{li-etal-2018-multi}. Besides, our method has better space utilization and integrate multiple designated subspaces information.

\subsection{Superiority of Dynamic Weights}

\begin{figure}[t]
\includegraphics[width=2.85in]{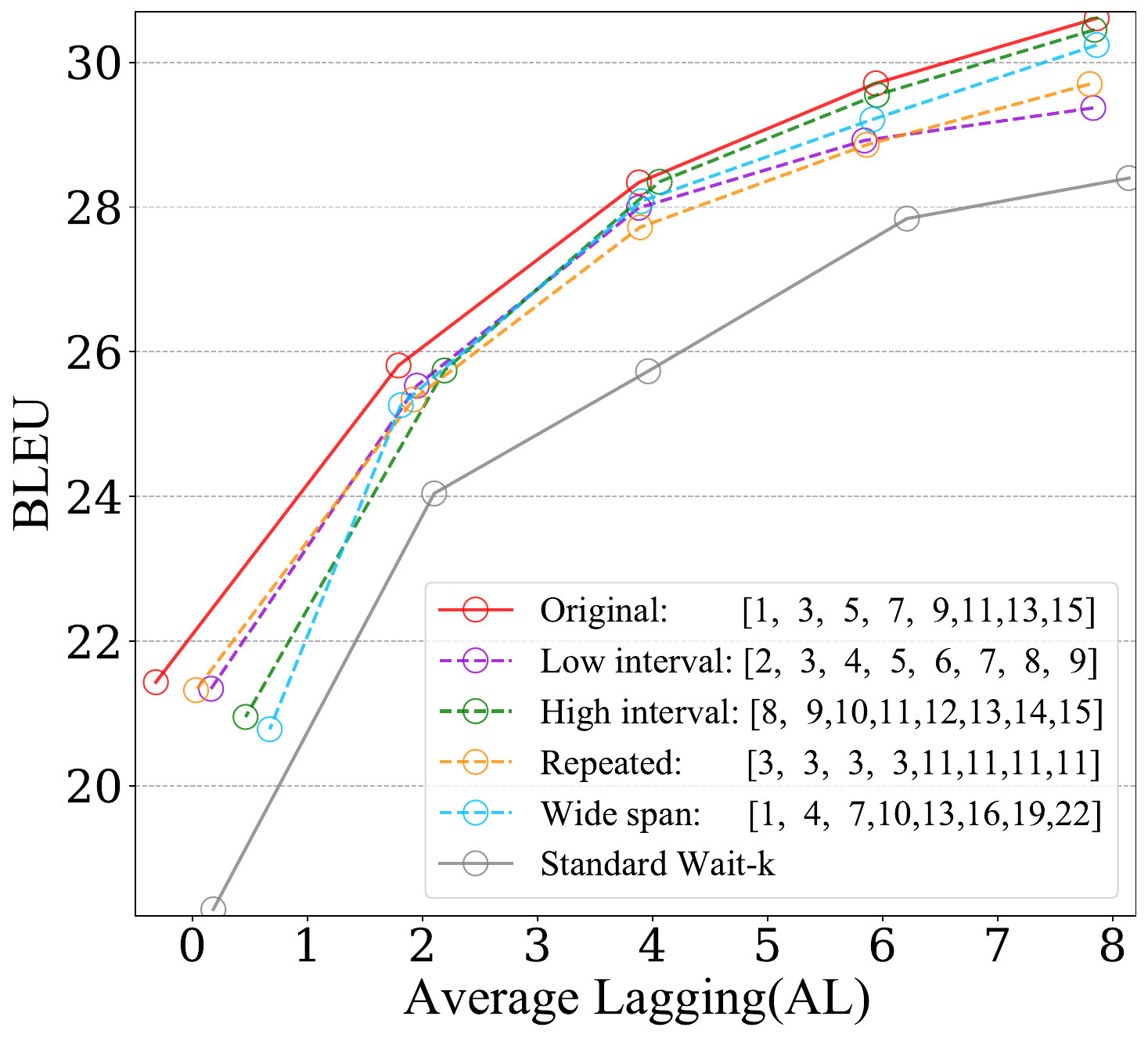}
\caption{Results of various settings of expert lagging $K_{\mathrm{\!MoE}}$ in MoE wait-k.}
\label{expert}
\end{figure}


Different expert outputs are dynamically weighted to achieve the best performance under the current test latency, so we calculated the average weight of each expert under different latency in Table \ref{gate2}.

Through dynamic weighting, the expert lagging of the expert with the highest weight is similar to the $k_{train}$ of the optimal model with standard wait-k, meanwhile avoiding the traversal on many trained models. When the test lagging is larger, the expert with larger expert lagging has higher weight; and vice versa. Besides, the expert with a slightly larger expert lagging than $k_{test}$ tends to get the highest weight for better translation, which is in line with the previous conclusions \cite{ma-etal-2019-stacl,future-guided}. Furthermore, our method enables the model to comprehensively consider various expert outputs with dynamic weights, thereby getting a more comprehensive translation.

\subsection{Effect of Expert Lagging}
\label{sec:lagging}

Expert lagging $K_{\mathrm{\!MoE}}$ is the hyperparameter we set to control the lagging of each expert. We experimented with several settings of $K_{\mathrm{\!MoE}}$ to study the effects of different expert lagging $K_{\mathrm{\!MoE}}$, as shown in Figure \ref{expert}.

Totally, all types of $K_{\mathrm{\!MoE}}$ outperform the baseline, and different $K_{\mathrm{\!MoE}}$ only has a slight impact on the performance, which shows that our method is not sensitive to how to set $K_{\mathrm{\!MoE}}$. Furthermore, there are some subtle differences between different $K_{\mathrm{\!MoE}}$, where the `Original' setting performs best. `Low interval' and `High interval' only perform well under a part of the latency, as their $K_{\mathrm{\!MoE}}$ is only concentrated in a small lagging interval. `Repeated' performs not well as the diversity of expert lagging is poor, which lost the advantages of MoE. The performance of `Wide span' drops under low latency, because the average length of the sentence is about 20 tokens where the much larger lagging is not conducive to low latency SiMT. 

In summary, we give a general method for setting expert lagging $K_{\mathrm{\!MoE}}$. $K_{\mathrm{\!MoE}}$ should maintain diversity and be uniformly distributed in a reasonable lagging interval, such as lagging 1 to 15 tokens.

\section{Conclusion and Future Work}
In this paper, we propose Mixture-of-Experts Wait-k Policy to develop a universal SiMT, which can perform high quality SiMT under arbitrary latency to fulfill different scenarios. Experiments and analyses show that our method achieves promising results on performance, efficiency and robustness.

In the future, since MoE wait-k develops a universal SiMT model with high quality, it can be applied as a SiMT kernel to cooperate with refined external policy, to further improve performance.

\section*{Acknowledgements}
We thank all the anonymous reviewers for their insightful and valuable comments. This work was supported by National Key R\&D Program of China (NO. 2017YFE0192900).

\bibliography{custom.bib}
\bibliographystyle{acl_natbib.bst}




\end{document}